\documentclass{article}

\usepackage{microtype}
\usepackage{graphicx}
\usepackage{subfigure}
\usepackage{booktabs} 

\usepackage{hyperref}

\usepackage[accepted]{icml2021}

\usepackage{booktabs}       
\usepackage{amsmath}
\usepackage{amssymb}
\usepackage{amsthm}
\usepackage{mathtools}
\usepackage{tcolorbox}
\usepackage{caption}
\usepackage{graphicx}
\usepackage{pdfpages}
\usepackage{float}
\usepackage{balance}

\usepackage{xcolor}
\definecolor{mygreen}{RGB}{102,252,102}

\theoremstyle{plain}
\newtheorem{theorem}{Theorem}

\numberwithin{lemma}{section}

\theoremstyle{definition}

\theoremstyle{remark}
\newtheorem{remark}{Remark}

\newcommand{\C}{\mathcal{C}}
\newcommand{\R}{\mathbb{R}}
\newcommand{\E}{\mathbb{E}}

\newcommand{\T}{\mathcal{T}}
\newcommand{\I}{\mathcal{I}}

\newcommand{\cov}{\textup{cov}}

\newcommand{\nn}{\textup{nn}}
\newcommand{\AdaFNN}{\textup{AdaFNN}}
\newcommand{\FPCA}{\textup{FPCA}}

\renewcommand{\S}{S}

\renewcommand{\P}{\mathbb{P}}
\renewcommand{\vec}{\textup{vec}}
\newcommand\independent{\protect\mathpalette{\protect\independenT}{\perp}}
\def\independenT#1#2{\mathrel{\rlap{$#1#2$}\mkern2mu{#1#2}}}

\newcommand{\papertitle}{Deep Learning for Functional Data Analysis with Adaptive Basis Layers}

\icmltitlerunning{\papertitle}

\defcitealias{CDCData}{NCHS, CDC 2020}

\begin{document}

\twocolumn[\icmltitle{\papertitle}



\icmlsetsymbol{equal}{*}

\begin{icmlauthorlist}
\icmlauthor{Junwen Yao}{af1}
\icmlauthor{Jonas Mueller}{af2}
\icmlauthor{Jane-Ling Wang}{af1}
\end{icmlauthorlist}

\icmlaffiliation{af1}{UC Davis}
\icmlaffiliation{af2}{Amazon (work done prior to joining Amazon)}

\icmlcorrespondingauthor{Junwen Yao}{jwyao@ucdavis.edu}

\icmlkeywords{Machine Learning, ICML}

\vskip 0.3in
]



\printAffiliationsAndNotice{}  

\begin{abstract}
Despite their widespread success, the application of deep neural networks to functional data remains scarce today. The infinite dimensionality of functional data means standard learning algorithms can be applied only after appropriate dimension reduction, typically achieved via basis expansions. Currently, these bases are  chosen a priori without the information for the task at hand and thus may not be effective for the designated task. We instead propose to adaptively learn these bases in an end-to-end fashion. We introduce neural networks that employ a new Basis Layer whose hidden units are each basis functions  themselves implemented as a micro neural network. Our architecture learns to apply parsimonious dimension reduction to functional inputs that focuses only on information relevant to the target rather than irrelevant variation in the input function. Across numerous classification/regression tasks with functional data, our method empirically outperforms other types of neural networks, and we prove that our approach is statistically consistent with low generalization error. Code is available at: \url{https://github.com/jwyyy/AdaFNN}.
\end{abstract}

\section{Introduction} \label{introduction}

Deep learning has revolutionized data analysis and predictive modeling as its learned input representations capture more relevant aspects of a problem than representations based on manually selected features of the data. While the powerful capabilities of neural networks (NN) have been clearly demonstrated for vector/image/text/audio/graph data, how to best adapt these models to functional data remains under-explored. Functional data are sample of random functions. The simplest examples are random curves defined over a  (univariate) real-valued interval with one curve per individual subject in our dataset. Without loss of generality we assume that the interval is $[0, 1]$. The curve for one subject is thus a random function $X(t), t \in [0, 1]$, which can be viewed as a continuous stochastic process on $[0, 1]$. Extensively studied in the statistics literature  \citep{FV2006, R2007, HE2015, W2016}, functional data appear frequently in scientific studies and daily life, such as in datasets of:  
air pollution, fMRI scans, growth curves, and sensors like wearable devices.   

A fundamental property that distinguishes functional data from other data types is that they are intrinsically infinite dimensional and generated by smooth underlying processes. While high-dimensionality poses challenges in modeling and prediction, it brings benefits when data are generated from smooth or continuous functions. This is because the observed measurements at one location $t_0$ can inform us the values of $X(t)$ for $t$  at nearby locations, thereby increasing the estimation efficiency. The smoothness assumption also makes functional data resilient to noise contamination as the magnitude of the noise can be estimated and statistical  methods designed for functional data can accommodate noise in the observed data.

In practical applications, we are interested in using these continuous curves $X(t)$ to infer their relationship to some response variable  $Y$, often to predict the response. 
More formally, we have a dataset $\{ (X_i(t), Y_i)\}_{i=1}^n$ of i.i.d. samples of $X(t)$ and the associated $Y$, where $X (t)$ is a mean-square continuous process defined over $t \in [0,1]$. For each subject $i$ in our dataset, the observations of $X_i$ serve as a functional covariate to predict scalar response  variable $Y_i$, which may be either continuous or discrete. 
In many practical applications, the continuous process $X(t)$ is only be observed  on a discrete time grid $\{t_1, \ldots, t_{J+1} \}$ in each observed $X_i$. 
Assuming  there is an underlying map $\T : X (t) \mapsto Y$, our goal is to estimate $\T$ from the data (e.g.\ using a neural network). 

A common pipeline for \emph{functional data analysis} (FDA) is to summarize the information contained in each function into a finite-dimensional vector and then carry out the analysis using existing models for the resulting multivariate data. Two popular dimension reduction approaches are Functional Principal Component Analysis (FPCA) \citep{BR1986,  RS1991, S1991, YMW2005a, LH2010} and preselected basis expansions  using, for example, B-splines \citep{RW2001, CFS1991}. 

Existing approaches to deep learning for functional data rely on a straightforward pipeline that first applies classic functional data analysis methods and then a neural network in sequence. \citet{RCF2002} handled functional data through discretization and functional parameterization of weight matrices, while \citet{RDCV2005} and \citet{G2016} use   basis function expansion to convert functional inputs into a vector form that can then be directly fed into to a standard  neural network. Some  universal approximation theory for functional neural networks has  been established by \citet{RDCV2005} and \citet{GS2019}. However this two-stage fitting process in prior work is unable to fully leverage the representation-learning power and flexibility of deep learning.

In this paper, we propose to improve existing architectures by replacing these pre-specified choice of basis functions with adaptively learned bases that are implemented via micro neural networks \citep{LCY2013} to which we backpropagate information regarding the response variable. A similar variant of our basic idea was briefly suggested by \citet{RC2005}, but their work never pursued the idea beyond a short comment stating it could be the possibile to implement the weight function of their functional neural network as an Multilayer Perceptron (MLP). 
Our design eliminates the need for a preprocessing step to convert random functions into vector inputs that otherwise typically requires a manually-prespecified choice of basis functions. Our architecture synchronizes the dimension reduction step and the nonlinear mapping step by adjusting the learned basis functions such that they only capture the information in $X(t)$ that is  relevant to the output. 
Existing basis function representations instead seek to retain as much information about the input as possible, which may actually make supervised learning more difficult \citep{tishby2015deep}.

Our main contribution is to propose an alternative neural architecture for \emph{end-to-end} FDA that consists of a novel \textit{Basis Layer} (BL) implemented via micro networks. We name the new network an \textit{Adaptive Functional Neural Network} (AdaFNN). We study some theoretical properties of AdaFNN, establishing convergence and generalization error guarantees.  Adding a BL into a neural network as feature extractors for functional inputs, the resulting model is empirically more accurate than existing methods and is simultaneously more parsimonious (meaning it requires less basis functions to model the random functions). Two types of regularizers are introduced to encourage basis orthogonality and basis sparsity. This regularization improves the resulting learned representations as well as the interpretability of the learned bases (especially when a small number of basis nodes are used). Moreover, our model can be trained end-to-end and thus composed with arbitrary differentiable operations without any alteration.

\section{Related Work} \label{related_work}

\subsection{Discretization of Functions}

A  straightforward application of neural networks to functional data is to treat the discretely observed functional values $\{X(t_1), \ldots, X(t_{J+1})\}$ as a high-dimensional vector and then input this vector into a neural network. This approach has been explored in \citet{RCF2002, RC2005, RDCV2005, GS2019}. An alternative common approach is to treat the discretely sampled data from  subjects as time series. However this approach has several disadvantages as  it does not leverage the smoothness of the underlying data-generating process $X(t)$ and relies on additional 
strong assumptions such as stationarity. Lastly, most time-series methods are not designed for replicate observations (here we observe numerous draws of the underlying $X(t)$, one per subject). Thus we instead review the vector-based approach below. 
Since  the $X(t)$ process is only observed at discrete time points $\{t_j\}_{j=1}^{J+1}$, one could use the vector $v_x = [ X (t_1), \dots, X (t_{J+1}) ]$ as the input of a standard neural network. Using the vector $v_x$, the original mapping $\T$ can be approximated by $\T_{\text{finite}}: v_x \mapsto Y$. Classical results imply that $\T_{\text{finite}}$ can be well approximated by a neural network with a sufficient number of parameters. Furthermore, by increasing the partition resolution $J$, we can  approximate $\T$ using $\T_{\text{finite}}$ with arbitrarily small error. 

\textbf{Drawbacks.} 
In order to preserve critical information about the  functional inputs, the discretization approach may require a high-dimensional vector, which hampers subsequent learning due to the curse of dimensionality. Furthermore, these discrete vector dimensions may fail to reflect the smoothness inherent to many functional covariates if they are contaminated by noise. 

\subsection{Basis Representation of Functions}

To overcome the disadvantage of discretization, \citet{RC2005} proposed to make use of the continuity of functional data   and find  a better finite-dimensional representation of a functional input before feeding it into a network. To be specific, let $\{ \varphi_k (t) \}_{k = 1}^K$ be a set of $K$ continuous basis functions defined on $[0,1]$. The task is to represent $X (t)$ using a vector $v_a = [a_1, \dots, a_K]$ such that: 
\small
\begin{equation} \label{eq:expansion}
    X (t) \approx \sum_{k = 1}^K a_k \varphi_k (t).
\end{equation}
\normalsize
Commonly used basis functions are Fourier basis functions, B-splines, or  eigenfunctions  obtained via spectral decomposition of $\cov (X(s), X(t))$. After finding a basis expansion of $X(t)$, we can use the vector $v_a$ instead of $v_x$ as the input to a feedforward network. Normally the dimension $K$ of $v_a$ is much smaller than the dimension $J+1$ 
of the discretized data $v_x$, a clear advantage.  Furthermore, the basis expansion in (\ref{eq:expansion}) automatically produces a smooth approximation of $X(t)$, which can reduce the noise contained in  $v_x$. 

\textbf{Drawbacks.} While the basis representation approach in (\ref{eq:expansion}) could  recover the underlying smooth process of functional input, it does not take advantage of the key information contained in the response $Y$ during its dimension reduction stage. Besides, both dimension reduction of functional inputs and parameterization of weight matrices (see FMLP on p.55, \citet{RC2005}) require selection of basis functions. These bases are typically  selected a priori and the number of bases  needs to be chosen as well. We address these questions in this paper by proposing an adaptive approach to find the optimal bases that utilizes the information on $Y$ and the specific learning task.

\subsection{Micro Network Inside of a Network}

Embedding smaller neural networks within a larger overall network architecture has been previously explored. For example, the Network in Network (NIN) model of \citet{LCY2013} replaces linear convolutions by a micro MLP. Empirically, the enhanced nonlinearity improves the model's ability to extract good features. Although NIN is conceptually related to our proposal, their operations differ in two ways. In our design, a basis node in a basis layer, which is also a micro MLP, is applied to the whole input. In contrast, a NIN micro network performs local convolution operations. Second, the micro MLP in NIN takes a small region of an image as its input and this process is slid across the whole image via convolution. Our basis micro network instead takes a fixed time point $t$ as its input and this process operates on a full functional input $[X(t_1),\dots,X(t_{J+1})]$ via numerical integration.

\section{Methodology} \label{methodology}

\begin{figure}[tb]
\vskip 0.2in
\begin{center}
\centerline{\includegraphics[width=0.78\linewidth]{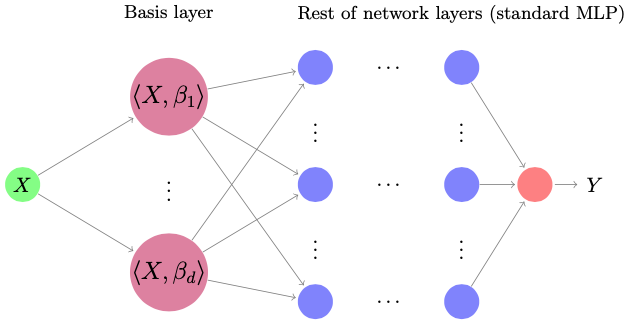}}
\caption{Neural network with our Basis Layer} 
\label{fnn}
\end{center}
\vskip -0.15in
\end{figure}

\begin{figure}[tb]
\begin{center}
\centerline{\includegraphics[width=0.95\linewidth]{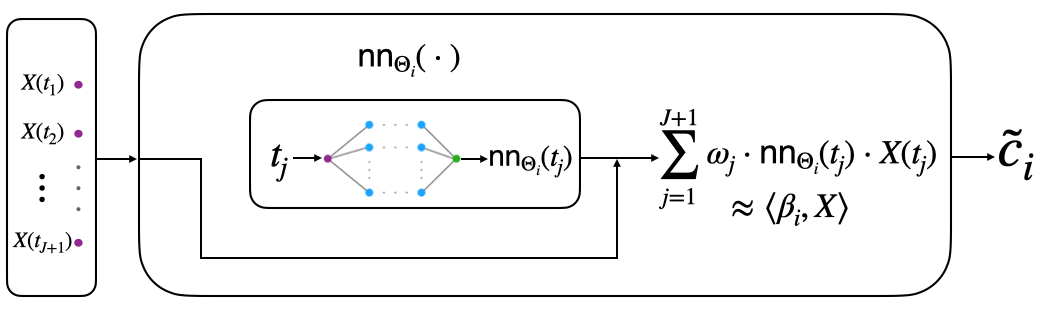}}
\vspace*{-0.5em}
\caption{The $i$-th basis node in a Basis Layer.} 
\label{nnBL}
\end{center}
\end{figure}

To address the drawbacks of discretization and basis representation, we propose a novel neural network that adaptively learn the best basis functions for supervised learning tasks with functional inputs. Figure~\ref{fnn} shows the basic architecture of such a network (AdaFNN), and Algorithm~\ref{AdaFNN_alg} details the network computations used to produce predictions in a forward pass. After the initial basis layer, our network shares the same structure as a standard feedforward network. Each node in a BL  outputs a scalar value computed as the inner product between the input function and the corresponding basis function at that node (Figure~\ref{nnBL}). Unlike handcrafted functions used in existing methods, we parameterize each basis function with a micro neural network that takes a scalar $t$ as its input and outputs the value the basis function takes at $t$. 

In the network depicted in Figure \ref{fnn}, the BL consists of $d$ basis nodes (for some user-specified value of $d$). Each node represents the application of some basis function $\beta_i (t)$, for $i = 1, \dots, d$ and $t \in [0,1]$. The value output by each basis node, i.e.,\ the \emph{score} of $X(t)$ with respect to the basis function $\beta_i (t)$, is computed as: 
\begin{equation}
c_i = \langle \beta_i, X \rangle = \int \beta_i(t) \cdot X(t) \ \mathrm{d}t.
\label{eq:integral} 
\end{equation}
\\[-1em]
The BL outputs form a vector $c = [c_1, \dots, c_d] \in \mathbb{R}^d$. This vector is then fed through the rest of the AdaFNN network's layers, which after the BL are all standard fully-connected layers (as in a standard MLP, where $\Theta$ denotes the weights of all layers after the BL). The output of the overall network is 
$\hat{Y} = \widehat\T (X) = \sigma_L \big( \cdots \sigma_1 \big( W_1 c + b_1 \big) \big)$, where $\sigma_1, \dots, \sigma_L$ are the activation functions at each layer.

The bases $\beta_i$ are used to project the functional input $X(t)$ to a vector representation $c$. While the form of these bases could in theory be selected a priori, such ad hoc selection violates the principle of end-to-end learning that drives deep learning's success. We would instead like to backpropagate information about $Y$ through the network in order to adaptively identify optimal bases $\beta_i$. For this reason, we prefer the representation of input functions in (\ref{eq:integral}) over (\ref{eq:expansion}), as the latter is less amenable to  basis learning via backpropagation.

In the BL, each basis function $\beta_i (t)$ is itself  parameterized by another neural network $\textup{nn}_{\Theta_i} (t) : t \mapsto \sigma^i_L \big( \cdots \sigma^i_1 \big( W^i_1 t + b^i_1 \big) \big)$ with parameters $\Theta_i$ consisting of weights $\{ W^i_l \}_{\ell=1}^L$ and biases $\{b^i_l\}_{\ell=1}^L$. In this work, these micro networks $\textup{nn}_{\Theta_i}$ are simply MLPs whose only input are the values $t_j$ at which the functional covariate $X(t)$ is measured (our micro MLPs employ modern techniques such as skip connections, batch/layer normalization, and dropout). Note that these are the only locations at which we need to evaluate the learned basis function $\beta_i$ in order to approximate $\langle \beta_i, X \rangle$. In practice, the integral in (\ref{eq:integral}) must be approximated numerically, for example, using the rectangular  or trapezoidal rule. Suppose that the domain $[0,1]$ is equally discretized by the partition $\{ t_j = (j - 1) / J \}_{j = 1}^{J+1}$. For each $i$, the score $c_i$ is approximated by $\tilde{c}_i = \sum_{j = 1}^{J+1} \omega_j \cdot \textup{nn}_{\Theta_i} (t_j) \cdot X (t_j)$, where $\{ \omega_j \}_{j = 1}^{J+1}$ are weights used in a numerical integration algorithm. For instance, if trapezoidal rule is used in the network, then we would have $\omega_{J+1} = \omega_1 = 1/(2J)$ and $\omega_j = 1/J$ for $j = 2, \dots, J$. Here  equal spacing is not necessary. Our method works well as long as the grid is dense enough for the numerical integration to be accurate (see our provided code for a demonstration of AdaFNN with non-uniform spacing). 

 The loss of the network is calculated between a prediction $\hat{Y}$ and the observed response $Y$, and is written as $\ell (\hat{Y}, Y)$. Here we employ a standard loss function for prediction such as mean-squared-error for regression or cross-entropy for classification. 
The overall loss is the average loss across the whole sample (or a mini-batch of data for stochastic gradient training) and is denoted as $L (\{ \Theta_i \}_{i=1}^d, \Theta) = \frac{1}{n} \sum_{i = 1}^n \ell (\hat{Y}_i, Y_i)$.  
The micro-networks and subsequent fully-connected network layers are all simultaneously trained (end-to-end) using this single objective $L(\{ \Theta_i \}_{i=1}^d, \Theta)$ applied to the output predictions. In the next section, we also consider possible addition of orthogonality or sparsity  regularization to this objective. 

\begin{algorithm}[tb]
   \caption{AdaFNN Forward Pass}
   \label{AdaFNN_alg}
\begin{algorithmic}
    \STATE {\bfseries Input:} data $x = [ X(t_1),\dots, (X_{J+1}) ]$
    \STATE {\bfseries Output:} prediction $\hat{y}$
    \STATE {\bfseries Parameters:} basis layer NN  $\{\Theta_i\}_{i = 1,...,d}$, output NN $\Theta$, 
    integration weights $\omega_1, \dots, \omega_{J+1} $, e.g.\ for trapezoid rule with equally-spaced $t_j$: \  $\omega_j = \frac{1}{J}$ if $2 \le j \le J$, else $=\frac{1}{2J}$ 
    \\[-0.2cm] \hrulefill
    \STATE \textbf{for} $i=1$ {\bfseries to} $d$: \ \ 
    $\tilde{c}_i \leftarrow \sum^{J+1}_{j = 1} \omega_j \cdot \nn_{\Theta_i} (t_j) \cdot X (t_j)$ 
    \STATE $\tilde{v}_c \leftarrow [\tilde{c}_1, \dots, \tilde{c}_d] \in \mathbb{R}^d$ 
    \STATE $\hat{y} \leftarrow \text{nn}_\Theta (\tilde{v}_c)$
    \STATE {\bfseries return} $\hat{y}$
\end{algorithmic}
\end{algorithm}

We make three observations about the proposed model. 
First, it can clearly be trained end-to-end with the BL. There is no need to select and even fine tune what type of basis functions should be used in representing input  functions (or weight matrices). Second, since the parameters $\Theta_i$ are updated to minimize prediction loss, our learned basis functions are likely better suited for the desired task than handcrafted basis functions chosen without this information. The experiments in Section \ref{experiments} show that our BL architecture can achieve higher accuracy than existing basis expansion methods while utilizing fewer basis functions (as each individual learned basis function can capture more predictive signal).
Lastly, since each micro neural network is a composition of continuous functions (as a standard MLP), all of the learned basis functions  within our model will be continuous.

\subsection{Regularization\label{regularization}} 

\paragraph{Encouraging Basis Orthogonality.} 

Without constraints, two BL nodes might learn similar basis functions and extract redundant information from the same functional input $X(t)$. To encourage different BL nodes to represent different (uncorrelated) information about the function, we can regularize them to be orthogonal. Recall that $L \big( \{ \Theta_i \}_{i =1}^d, \Theta \big)$ is the loss function of a BL network. To encourage basis orthogonality, we introduce a regularization term which penalizes the cosine similarity between each pair of basis functions. The resulting loss optimized by the regularized network is  \\[-2.5em]

\small
\begin{align*}
& L_{\text{perp}} \big( \{ \Theta_i \}_{i =1}^d, \Theta \big) \\
& = L \big( \{ \Theta_i \}_{i =1}^d, \Theta \big) + \lambda_1 \cdot \frac{1}{\binom{d}{2}} \sum_{j \ne j'} \frac{|\langle \nn_{\Theta_j}, \nn_{\Theta_{j'}} \rangle |}{\| \nn_{\Theta_j} \|_2 \| \nn_{\Theta_{j'}} \|_2} , 
\end{align*} \\[-2.5em]

\normalsize
where $\lambda_1 > 0$ controls the strength of the penalty. When a BL contains many nodes, enumerating all pairs becomes computationally expensive. Instead we randomly sample a few pairs at each mini-batch update and employ their average absolute cosine similarity as a stochastic estimate of the regularizer against which we optimize network parameters.

\paragraph{Encouraging Basis Sparsity.} 
We can also regularize the shape of our learned basis functions. In domain selection problems \citep{JWZ09, ZWW13, WLZH21},  the response is only related to the functional input over an (a priori unknown) subset of its domain, i.e.\ ${Y \independent \{X(t')\}_{t' \notin \I } \mid \{X(t)\}_{t \in \I }}$ for some $\I \subset [0,1]$. 
To encourage this desired property, it is sensible to learn a basis function whose value is zero outside of $\I$. While the number of nonzero values taken by the basis function is hard to optimize, we can penalize their $L_1$ norm as a tight convex relaxation of an $L_0$ norm to enforce basis sparsity. The resulted loss function is \\[-1em]
\begin{align*}
L_{\textup{sprs}} \big( \{ \Theta_i \}_{i =1}^d, \hspace*{-0.5mm} \Theta \big)  = L \big( \{ \Theta_i \}_{i =1}^d,  \hspace*{-0.5mm} \Theta \big) + \lambda_2 \cdot \frac{1}{s}  \hspace*{-0.5mm} \sum_{i \in \mathcal{S}}  \hspace*{-0.1mm} \int  \hspace*{-1mm} |\beta_i (t)|  \hspace*{-0.5mm} \ \mathrm{d}t, 
\end{align*} \\[-1em]
where $\mathcal{S} \subseteq \{1,\dots,d\}$ indicates which subset of basis functions we wish to sparsify, $s$ is the number of elements in $\mathcal{S}$, and $\lambda_2 > 0$ controls the strength of the $L_1$ penalty. Even when we are not sure whether the domain selection assumption truly hold, learning sparse basis functions via this $L_1$ penalty can greatly improve the overall interpretability of our prediction model (Figure \ref{plot_reg}).

\section{Theoretical Analysis} \label{theory}
Here we discuss theoretical properties of the architecture proposed in the previous section. We provide a universal approximation theorem for this design and prove it achieves low generalization error under mild regularity conditions.

Let $\C([0,1])$ denote the space of continuous functions defined on the compact interval $[0,1]$. Assume that the underlying mapping $\T : X \mapsto Y$ is a composite of a finite-dimensional linear transformation and a subsequent non-linear transformation. We can write $\T = h \circ g$, where $g : \C ([0,1]) \to \R^q$ is a linear continuous map, and $h : \R^q \to \R$ is a non-linear continuous map. By Riesz representation theorem, there exist square-integrable function $\gamma_i$ with $i=1,\dots,q$ such that $g(X) = [\langle \gamma_1, X\rangle, \dots, \langle \gamma_q, X \rangle]$. The approximate network parameterized by weights $\{\Theta_i\}_{i=1}^q$ and $\Theta$ is denoted as $\widehat\T$.

\begin{theorem}[Consistency of the network] \label{theorem1}
With the notations defined previously and following the conventions in the literature, we assume that: 
\\[-2em]
\begin{enumerate}
\item [(i)] the numerical integration at each basis node can be accurately evaluated as $J \to \infty$,  
\\[-2em]
\item [(ii)] each network in $\widehat{\T}$ can have sufficient capacity.
\\[-2em]
\end{enumerate}
Then for any $\epsilon > 0$, there exists a network $\widehat{\T}^\ast$ with weights $\{ \Theta_i^\ast \}_{i = 1}^q$ and $\Theta^\ast$ such that 
\begin{align*} 
\sup_{f \in \C ([0,1]), \| f \|_2 \leq 1} | 
\widehat{\T}^\ast(f) - \T (f) | < \epsilon .
\end{align*} 
Hence, we have the following result: Let $X$ be a continuous process defined on $[0,1]$. For any $\delta > 0$, there exists a network $\widehat{\T}^\star$ with weights $\{ \Theta_i^\star \}_{i = 1}^q$ and $\Theta^\star$ such that 
\[ \P \big( | \widehat{\T}^\star(X) - \T (X) | < \delta \big) > 1 - \delta .\]
\end{theorem}
\begin{remark} \label{remark1}
Although Theorem \ref{theorem1} provides the consistency of the proposed architecture, it is not equivalent to identifying each true basis function consistently. The reason is that the individual basis functions are not identifiable since there are multiple ways to parameterize one map. 
\end{remark}

\begin{remark} \label{remark2} 
To make adequate predictions, the number of basis nodes $d$ in AdaFNN should be sufficiently larger than the dimensionality $q$ of $g(X)$,  introduced in the assumptions of Theorem 1. 
In practice, we could also vary the choice of the number of basis nodes to find out which yields the best performance (lowest validation loss or fewest bases with low validation loss). 
Once the training is done, by investigating the learned bases, one can decide which ones seem to be relevant and use those as the basis functions to re-train a smaller subsequent network. 
\end{remark}

Next, we prove the proposed architecture can achieve small generalization error. Let $\S = \{ (X_1, Y_1), \dots, (X_n, Y_n) \}$ be $n$ i.i.d. copies of $(X,Y)$, and there exist two constants $M_1, M_2 > 0$ such that both $\sup_{t\in[0,1]} |X (t)| \leq M_1$ and $|Y| \leq M_2$ hold almost surely.  Let $\widehat{\T}_\Theta$ be a model proposed in Section \ref{methodology} with its architecture fixed. In a slight abuse of notation, we use $\Theta$ to denote all the weights, including $\{\Theta_i\}_{i=1}^d$ and $\Theta$, used in $\widehat{\T}_\Theta$. We use $\ell \big( \widehat{\T}_\Theta (X) , Y \big)$ to denote the loss function. 
The population risk is defined as $r (\Theta) = \E \big[\ell \big( \widehat{\T}_\Theta (X) , Y \big)\big]$, and the empirical risk is $r_n (\Theta) = \frac{1}{n} \sum_{i = 1}^n \ell \big( \widehat{\T}_\Theta (X_i) , Y_i \big)$. Usually the weights $\Theta$ are estimated via $\widehat{\Theta}$ produced by a random algorithm $A$ based on the sample $\S$. 
The generalization error is defined as $\big| \E_{\S,A}[ r (\widehat{\Theta}) - r_n (\widehat{\Theta}) ] \big|$. Suppose that we use a variant of stochastic gradient descent to train our model. At the $t$-th iteration ($1 \leq t \leq T$), the weights are updated with  $\widehat{\Theta}_{t+1} = \widehat{\Theta}_t - \alpha_t \nabla_\Theta \ell \big( \widehat{\T}_{\Theta}(X_{i_t}), Y_{i_t} \big) \big|_{\Theta = \widehat\Theta_t}$, where the indices $i_t$ are randomly chosen (e.g., uniformly), and the learning rate $\alpha_t$ is monotonically non-increasing with $\alpha_t \leq c / t$ for some fixed constant $c > 0$. The following  result is an application of Theorem 3.12 in \citet{H2016}.

\begin{theorem}[Small generalization error] \label{theorem2}
Assume that
\\[-2em]
\begin{enumerate}
\item[(i)] the weight $\Theta$ is restricted on some compact region, 
\\[-2em]
\item[(ii)] the loss function $\ell(\cdot,\cdot)$ and its gradient $\nabla \ell(\cdot,\cdot)$ are both Lipschitz.
\end{enumerate} 
Then for every pair of observation $(X,Y)$, both $\ell \big( \widehat\T_\Theta(X), Y \big)$ and $\nabla_\Theta \ell \big( \widehat\T_\Theta(X), Y\big)$ are Lipschitz with respect to $\Theta$. Hence, there exists some constant $c > 1$ such that
\[  \big| \E_{\S,A}[ r (\widehat{\Theta}) - r_n (\widehat{\Theta})  ] \big| \lesssim \frac{T^{1 - 1/c}}{n}.\]
\end{theorem}

\begin{remark} \label{remark3}
Assumption (i) in Theorem \ref{theorem2} is not restrictive  since we can convert a constraint optimization problem to an equivalent penalized problem. In practice, we could minimize the regularized empirical risk, i.e., $r_n (\Theta) + \rho \| \vec (\Theta) \|_2$, where $\vec (\Theta)$ is the vectorized weights $\Theta$, and $\rho > 0$ is a tuning parameter.
\end{remark}

\section{Experiments} \label{experiments}

Throughout, our experiments focus on methods that can handle general functional inputs, rather than approaches tailored for specific tasks like predicting future observations. 
The proposed AdaFNN network is compared to three baseline models: `Raw data (\#) + NN' \citep{RCF2002}, `B-spline (\#) + NN' \citep{RDCV2005}, and `$\text{FPCA}_{\text{p}}$ + NN'\citep{RDCV2005}. Here the integer \# denotes the input dimension of each network. The subscript $p$ of `FPCA' is the fraction of variation explained (FVE) used in selecting the number of principal components. The latter two baselines (B-spline and FPCA) have their roots in the field of functional data analysis \citep{CFS1999,CFS2003,MS2005,DPZ2012} and classification tasks \citep{M2005, LM2006, SDK2008, C2012}.

The first baseline simply discretizes the raw functional input as a vector that is fed into the network, so the dimension of the input vector is  the number of points $t$ at which $X(t)$ has been observed. The other two  baseline models consist of two steps. The first step involves transforming the functional input into  a vector of scores  from its B-spline or  FPCA expansion (cf. (\ref{eq:expansion})).  The resulting vector representation is then fed as input into a network in the second step.

The number of bases used in the  two baseline models is often determined by how well a function can be represented in the functional space spanned by these basis functions. For example, when using B-splines, we look at how well   the selected spline functions can capture the trend of the raw data.  A small number of B-splines may not be able to recover the trajectory very well. However, too many B-splines might overfit  the functions. The choice of the number of principal components is  based on the desired FVE by these principal components. Often, a sizeable FVE is expected, e.g., 90\% to 99\%. Similar to the selection of B-splines, we should not simply choose an FVE as large as possible. Large FVE may include (functional) principal components whose corresponding eigenvalues are very small, hence difficult to estimate. A good rule of thumb  is to use the first few components that capture the bulk of the variation.

We write `AdaFNN ($\lambda_1$,$\lambda_2$)' to indicate the level of regularization, where $\lambda_1$ (and $\lambda_2$) controls the degree of orthogonality (and $L_1$) regularization. 
AdaFNN was trained with 9 different combinations of the orthogonal regularization penalty ${\lambda_1 \in \{0, 0.5, 1\}}$ and $L_1$ regularization penalty ${\lambda_2 \in \{0, 1, 2\}}$. The performance of each configuration is reported for all simulations and real data tasks. Throughout we use * to indicate the $\lambda_1, \lambda_2$ values that performed best on the validation data, as these are the hyperparameter values that would be typically used in practice.

All models, including AdaFNN and the other NN baselines, employ the same architecture and training hyperparameters. A network with 3 hidden layers and 128 fully connected nodes per layer was used for Tasks 1-7 (real data) and our simulation studies. For Tasks 8 and 9 with small sample sizes,  we used a smaller network with 2 hidden layers and 64 nodes and added dropout during training. All networks were trained up to 500 epochs (with 200-epoch early stopping patience) using mini-batches of size 128. AdaFNN merely uses 2 bases in simulation Cases 1 \& 4, 3 bases in simulation Cases 2 \& 3, and 4 bases in our applications to all 9 prediction tasks with real data. In contrast, we allowed the B-spline/FPCA baselines to either rely on a similar number or  more basis functions since these models cannot optimize each of their bases (e.g.\ $\text{FPCA}_{0.99}$ often selected more than 10 principal components).

\subsection{Simulation Studies}

We demonstrate through simulations that baseline functional neural network models may miss relevant information but AdaFNN is able to capture the true signal while relying on fewer basis functions than the baselines. Four different simulation settings were considered, each is purposefully designed to illustrate a particular conceptual shortcoming of one of the baseline methods (with an extra fifth setting to highlight the utility of our proposed regularization). 

We first describe the underlying data-generating process in each of the four  settings. For $t \in [0,1]$, define $\phi_1 (t) = 1$ and $\phi_k (t) = \sqrt{2} \cos ((k-1) \pi t)$, $k = 2,\dots,50$. Consider the process  $X (t) = \sum_{k = 1}^{50} c_k \phi_k (t)$, where $c_k = z_k r_k$, and $r_k$ are i.i.d. uniform random variables on $[-\sqrt{3}, \sqrt{3}]$. The actual observations for $X_i(t)$ is the discrete data $\{X_i(t_j), j=1, \dots, 51\}$. 
We report the \emph{mean squared prediction error} (MSE) achieved by each method on the test data.

\noindent {\bf Case 1:} We set:  $z_1 = 20, z_2 = z_3 = 5$, and $z_k = 1$ for $k \geq 4$. The response is $Y = c_3^2$, that is, $Y = (\langle X, \phi_3 \rangle)^2$, where the function $\phi_3$ corresponds to  the true predictive \emph{signal}. This case is designed to show that a small FVE in FPCA may not suffice to capture the relevant signal.

\noindent {\bf Case 2:} We set:  $z_1 = z_3 = 5, z_5 = z_{10} = 3$, and $z_k = 1$ for other $k$. The response is  $Y = c_5^2 = (\langle X, \phi_5 \rangle)^2$. Here the function $\phi_5$ corresponds to  the true predictive \emph{signal} and is more complex than $\phi_3$. The squared operation ensures a nonlinear relationship between the response and functional covariate. This case is designed to show that a small number of B-splines may not suffice to represent the $X(t)$ information relevant to the response.

\noindent {\bf Case 3:} In Cases 1 \& 2, both the response $Y$ is free of noise and the functional input $X(t)$ is not contaminated by measurement errors. 
However, this setup is rarely realistic in practice. The goal here is to evaluate  functional estimators in the presence of both outcome noise and measurement errors in $X(t)$. We use  the same  model as Case 2 except that a mean zero Gaussian noise is added to the response, and the observation of $X(t)$ at each time point is perturbed by an additive mean zero Gaussian measurement error. The signal-to-noise ratios (SNR), defined as 
\[ \small{ \text{SNR} (X) = \frac{\sqrt{\int_0^1 (X(t))^2 \ \mathrm{d} t}}{\text{standard deviation of  measurement error}} } \] 
for the functional input (due to $\E[X(t)] = 0$), is $\sqrt{10}$ to 1.

\noindent {\bf Case 4:}
This case studies how well AdaFNN captures multiple signals and its application in domain selection. The functional covariates $X_i(t)$ are  generated similarly as in Cases 1 \& 2, except that in Case 4: $z_i$ are all taken to be 1. Two signals $\beta_1$ and $\beta_2$ are chosen as: $\beta_1 (t) = (4 - 16t) \cdot 1\{0 \leq t \leq 1/4 \} $ and $\beta_2 (t) = (4 - 16|1/2-t|) \cdot 1\{1/4 \leq t \leq 3/4 \}$. The response is $Y = \langle \beta_2, X \rangle + (\langle \beta_1, X \rangle)^2$. Centered Gaussian noise is added to $Y$, and $X(t)$ is also contaminated by measurement error. 

\noindent{\bf Case 5:} 
The same setup as Case 4, but now with double the noise variance in $Y$ (used to highlight our regularization).

\begin{table}[ht]
\captionof{table}{Test-set MSE of predictions in simulation study. For each case, the asterisk indicates the best  AdaFNN hyperparameters on the validation set, and the method with the best test MSE is marked in bold.}
\label{table1}
\vspace*{-0.5em}
\begin{center}
\begin{small}
\begin{sc}
\begin{tabular}{l c @{\hskip 2mm} c @{\hskip 2mm} c @{\hskip 2mm} c @{\hskip 2mm} c } 
\toprule
Method & Case 1 & Case 2 & Case 3 & Case 4 \\ 
\cmidrule{1 - 5}
Raw data (51) + NN                    & 0.015 & 0.038 & 0.275 & 0.334  \\
B-spline (4) + NN                 & 0.050  & 0.984 & 0.971 & 0.369  \\
B-spline (15) + NN               & 0.013 & 0.019 & 0.206 &  0.251 \\
$\text{FPCA}_{0.9}$ + NN   & 0.917 & 0.023 & 0.134 & 0.855  \\
$\text{FPCA}_{0.99}$ + NN & 0.003 & 0.036 & 0.239 & 0.667  \\
\cmidrule{1-1}
AdaFNN (0.0, 0.0) & $\textbf{0.001}^*$ & \textbf{0.003} & 0.979 & $\textbf{0.193}^*$  \\
AdaFNN (0.0, 1.0) & 0.995 & 0.007 & 0.978  & 0.982 \\
AdaFNN (0.0, 2.0) & 0.996 & 0.992 & 0.978 & 0.981  \\
AdaFNN (0.5, 0.0) & 0.004 & $\text{ }0.005^*$ & $\text{ }0.137^*$ & 0.571  \\
AdaFNN (0.5, 1.0) & 0.983 & 0.005 & 0.978 & 0.590 \\
AdaFNN (0.5, 2.0) & 0.134 & 0.008 & 0.978 & 0.981  \\
AdaFNN (1.0, 0.0) & 1.000 & 0.004 & \textbf{0.127} & 0.196  \\
AdaFNN (1.0, 1.0) & 0.009 & 0.006 & 0.974 & 0.606 \\
AdaFNN (1.0, 2.0) & 0.051 & 0.009 & 0.978 & 0.981  \\
\bottomrule
\end{tabular}
\end{sc}
\end{small}
\end{center}
\vskip -0.1in
\end{table}

\begin{figure}[ht]
\begin{center}
\includegraphics[width=0.45\linewidth]{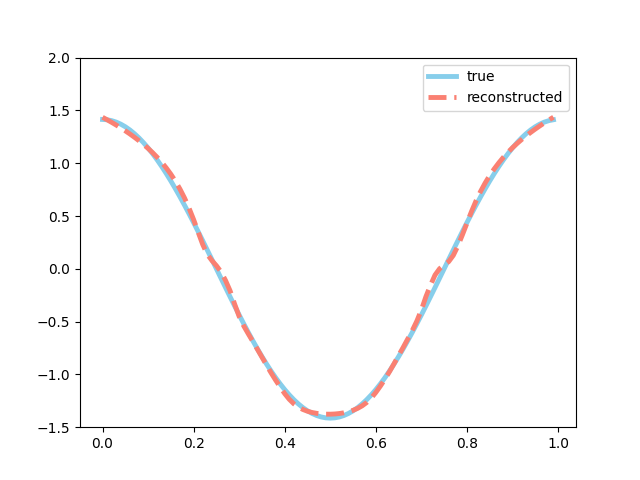}
\hfill
\includegraphics[width=0.45\linewidth]{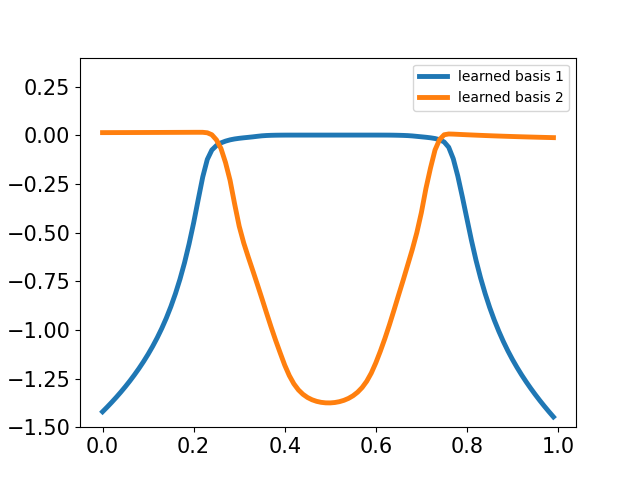} 
\vspace*{-0.5em}
\caption{\small 
The left plot shows the true signal $\phi_3$ (solid) and the reconstructed signal $\hat\phi_3$ (dashed) from $\AdaFNN(0,0)$ (the regularization values with best validation MSE) in Case 1. The right plot shows each learned bases $\hat\beta_1$ and $\hat\beta_2$ from the same experiment. Note that: $\phi_3\approx \hat\phi_3 = \hat\beta_2 - \hat\beta_1$. 
}
\label{sim1} 
\end{center}
\vskip -0.2in
\end{figure}

\begin{figure}[t]
\begin{center}
\includegraphics[width=0.45\linewidth]{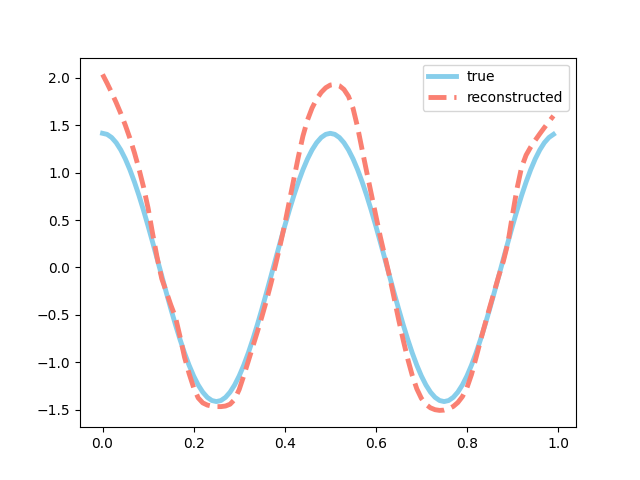} 
\hfill 
\includegraphics[width=0.45\linewidth]{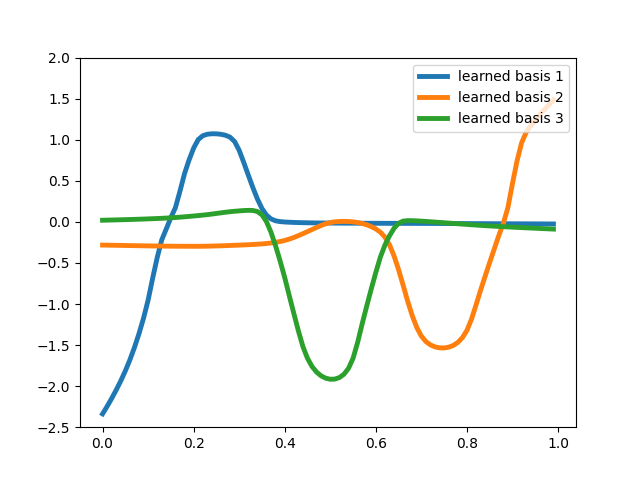} 
\vspace*{-0.5em}
\caption{\small 
The left plot shows the true signal $\phi_5$ (solid) and the reconstructed signal $\hat\phi_5$ (dashed) from $\AdaFNN(0.5,0)$ (the regularization values with best validation MSE) in Case 2. The right plot shows each learned bases $\hat\beta_1, \hat\beta_2, \hat\beta_3$ from the same experiment. Note that: $\phi_5\approx \hat\phi_5 =  \hat\beta_3 - \hat\beta_1 - \hat\beta_2$. 
} 
\label{sim2}
\end{center}
\vskip -0.2in
\end{figure}

Table~\ref{table1} reports the MSEs for all methods. Under columns Cases 1 \& 2, we see that AdaFNN exhibits the best performance in both settings. The performance of the baseline models  is mixed. The top two principal components  (with FVE at least 90\%) are not able to detect any useful predictive signal in the data under Case 1, same with four B-splines bases  in the simulation under Case 2. 
Thanks to its success in capturing the true basis function, AdaFNN performs strongly in 
both simulation settings. It is interesting to observe that each fitted basis function contains a fraction of the true signal function, and together they are able to recover it (Figures \ref{sim1} and \ref{sim2}). That is, the true signal function can be represented as a linear combination of the fitted bases.

For Case 3, Table~\ref{table1} (column `Case 3') shows that all methods performed worse with noise added, but AdaFNN with orthogonality regularization remains superior to other methods. On the other hand, the $L_1$ regularizer is not helpful in Case 3. This is expected, because the true signal $\phi_5$ does not have any zero region in its domain. Note that feeding the raw data as a vector into neural networks performs comparably to the two functional baseline models in the noiseless simulations. However, in noisy settings (Cases 3 \& 4), this approach is inferior to the other two baseline models. This demonstrates the importance of exploiting functional properties whenever the input is a smooth function. The performance of the functional `B-spline + NN' and `FPCA + NN' approaches is mixed; each has its own advantage over the other in different scenarios. 

\begin{figure}[t]
\begin{center}
\includegraphics[width=0.45\linewidth]{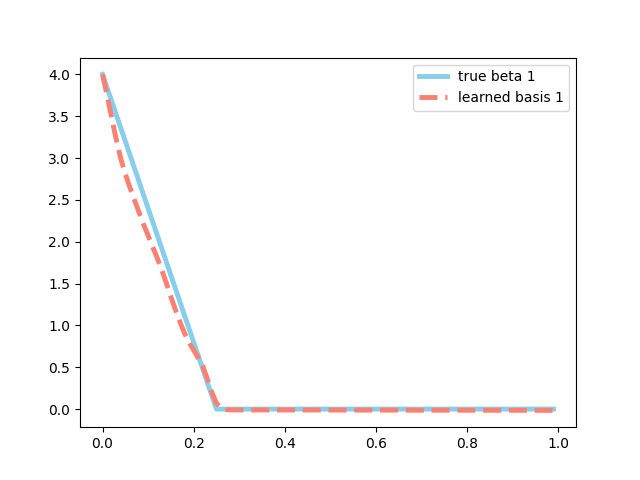} 
\hfill
\includegraphics[width=0.45\linewidth]{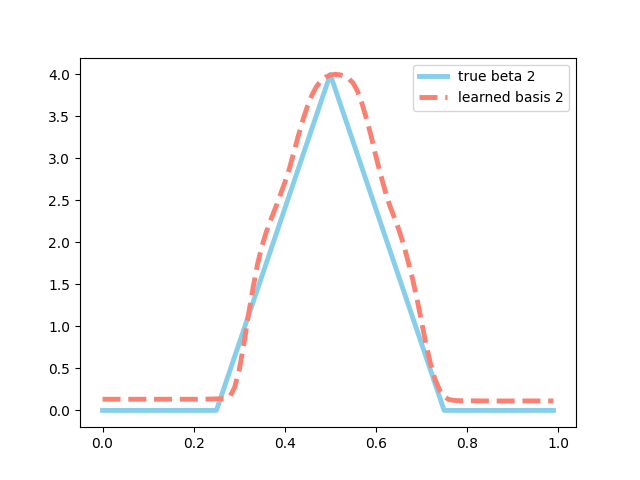} 
\vspace*{-0.5em}
\caption{\small 
The left plot shows the true signal $\beta_1$ (solid) and a (scaled) learned signal $\hat\beta_1$ (dashed) from $\AdaFNN(0,0)$ (the regularization values with best validation MSE) in Case 4. The right plot shows $\beta_2$ and a (scaled) $\hat\beta_2$ from the same experiment.
}
\label{sim4}
\end{center}
\vskip -0.2in
\end{figure}

For Case 4, Table~\ref{table1} (column `Case 4') shows that AdaFNN outperforms the other methods in this setting. At the same time, the proposed method also learns meaningful bases that correctly identify the relevant domain of interest  (Figure~\ref{sim4}). That on $[3/4, 1]$ both $\beta_1$ and $\beta_2$ are zero implies that the values of $X(t)$ over $[3/4, 1]$ have no effect on the response. None of the baseline methods is able to show this information, and thus AdaFNN is not only more accurate than existing models, but also more \emph{interpretable}. In summary, while some baseline methods perform well in certain simulation settings, none of these methods can achieve consistently strong performance like AdaFNN across all cases.

\begin{figure}[h]
\centering
\includegraphics[width=0.32\linewidth]{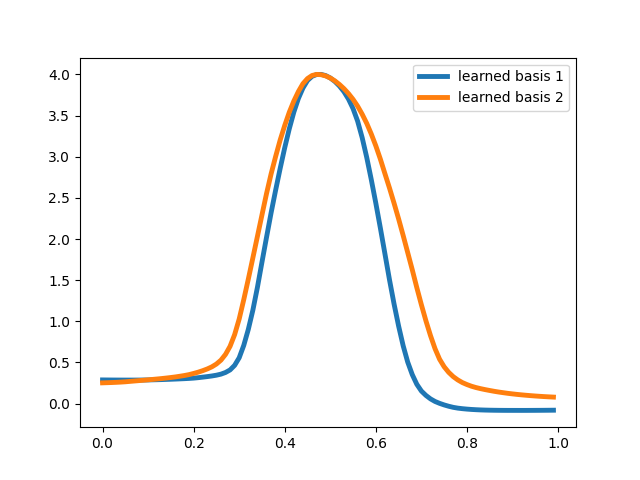}
\hfill
\includegraphics[width=0.32\linewidth]{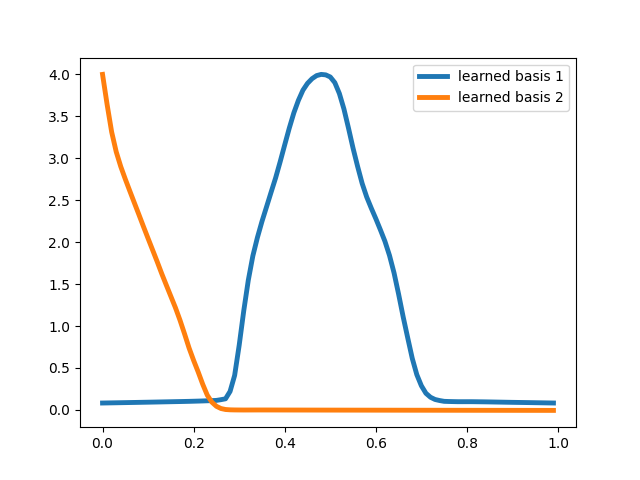}
\hfill
\includegraphics[width=0.32\linewidth]{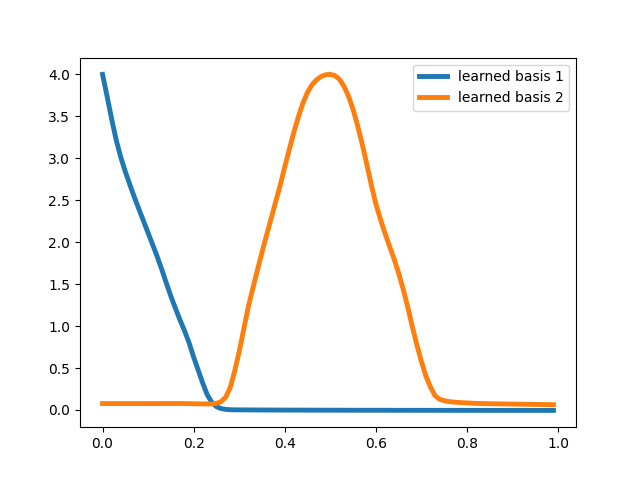}
\caption{
The (scaled) bases under simulation Case 5 learned by: $\AdaFNN(0,0)$ on the left,  $\AdaFNN(0.5,0)$ in the middle, and $\AdaFNN(0,0.1)$ on the right.
}
\label{plot_reg}
\end{figure}

A final simulation, Case 5, demonstrates the utility of our proposed regularizationn.  In this case, Table~\ref{table3} shows that $\AdaFNN(0.5,0)$ and $\AdaFNN(0,0.1)$ clearly outperform AdaFNN without regularization, as well as all other methods (despite our regularized AdaFNN using \textbf{only 2 bases}, under which the other methods performed very poorly). Without any regularization, $\AdaFNN(0,0)$ learns \textbf{2 very similar bases} (left plot in Figure~\ref{plot_reg}), while using either one of our proposed regularizers helps AdaFNN recover the true underlying bases (middle/right plots in Figure~\ref{plot_reg}) and greatly improves its predictive performance. 
\begin{table}[h]
\captionof{table}{Test-set MSE of predictions in Case 5. $\AdaFNN$ with active regularization is highlighted in bold.}
\label{table3}
\begin{center}
\begin{sc}
\begin{tabular}{l c @{\hskip 2mm} c @{\hskip 2mm} c }
\toprule
method & no. bases & MSE \\ 
\cmidrule{1 - 3}
Raw (51) + NN & 51  & 0.339 \\ 
B-spline (4) + NN & 4 & 0.382 \\ 
B-spline (15) + NN & 15 & 0.257 \\
$\FPCA_{0.9}$ + NN & 20 & 0.807 \\ 
$\FPCA_{0.99}$ NN & 28 & 0.693 \\ 
\cmidrule{1 - 1}
$\AdaFNN(0.0,0.0)$& 2 & 0.598 \\
$\AdaFNN(0.5,0.0)$& 2 & \textbf{0.231} \\
$\AdaFNN(0.0,0.1)$& 2 & \textbf{0.207} \\
\bottomrule
\end{tabular}
\end{sc}
\end{center}
\end{table}

\subsection{Application to Real Functional Datasets}

\begin{table*}[t]
\captionof{table}{Comparing test-set performance of different methods' predictions on 9 functional datasets (MSE in regression, $1-\textup{AUC}$ in classification). For each dataset, the asterisk indicates which AdaFNN hyperparameters performed best on the validation set, and the best performing method on the test data is indicated in bold.}
\label{table2}
\vspace*{-0.5em}
\begin{center}
\begin{small}
\begin{sc}
\begin{tabular}{l l cl cl cl cl cl cl cl cl cl} 
\toprule
Method & Task 1 & Task 2 & Task 3 & Task 4 & Task 5 & Task 6 & Task 7 & Task 8 & Task 9  \\ 
\cmidrule{1 - 10}
Raw data (48) + NN                    & 0.099 & 0.284 & 0.124 & 0.296 & 0.380 & 0.488 & 0.472 & 0.406 & 0.373 \\
B-spline (15) + NN               & 0.094 & 0.306 & 0.137 & 0.326 & 0.335 & \textbf{0.477} & 0.429 & 0.413 & 0.387 \\
$\text{FPCA}_{0.99}$ + NN        & 0.119 & 0.339 & 0.143 & 0.306 & 0.363 & 0.493 & 0.431 & 0.429 & 0.378 \\
\cmidrule{1-1}
AdaFNN (0.0, 0.0) & $\textbf{0.084}^*$ & $\text{ }0.290^*$ & $0.129^*$ & 0.311 & 0.365 & \textbf{0.477} & $\textbf{0.410}^*$ & $\text{ }0.377^*$ & 0.375 \\
AdaFNN (0.0, 1.0) & 0.094 & 0.276 & 0.126 & 0.327 & 0.561 & $\text{ }0.479^*$ & 0.498 & 0.374 & 0.392 \\
AdaFNN (0.0, 2.0) & 0.097 & 0.276 & 0.129 & 0.324 & 0.596 & 0.481 & 0.473 & 0.381 & 0.445 \\
AdaFNN (0.5, 0.0) & 0.108 & \textbf{0.260} & 0.130 & $\text{ }0.310^*$ & $0.380^*$ & 0.490 & \textbf{0.410} & 0.376 & $\textbf{0.368}^*$ \\
AdaFNN (0.5, 1.0) & 0.089 & 0.279 & 0.126 & 0.324 & 0.616 & 0.486 & 0.494 & \textbf{0.362} & 0.413 \\
AdaFNN (0.5, 2.0) & 0.098 & 0.280 & 0.128 & 0.345 & 0.392 & 0.509 & 0.444 & 0.373 & 0.450 \\
AdaFNN (1.0, 0.0) & \textbf{0.084} & 0.288 & \textbf{0.118} & \textbf{0.294} & \textbf{0.339} & 0.485 & 0.413 & 0.378 & 0.406 \\
AdaFNN (1.0, 1.0) & 0.097 & 0.282 & 0.133 & 0.320 & 0.651 & 0.502 & 0.456 & 0.371 & 0.394 \\
AdaFNN (1.0, 2.0) & 0.092 & 0.279 & 0.127 & 0.326 & 0.371 & 0.510 & 0.414 & 0.374 & 0.416 \\
\bottomrule
\end{tabular}
\end{sc}
\end{small}
\end{center}
\vskip -0.1in
\end{table*}

Next, we evaluate the performance of AdaFNN and other neural FDA methods in nine different regression and classification tasks, using four datasets. In regression tasks, the performance is again measured by MSE, while the performance is measured by the \emph{area under the ROC curve} (ROC  AUC) in classification tasks. Since our simulations show that B-spline (4) and $\text{FPCA}_{0.9}$ empirically underperform (Table~\ref{table1}), we subsequently only consider the use of B-spline (15) and $\text{FPCA}_{0.99}$ on real data.  

\noindent {\bf Electricity Data:} Electricity consumption readings for 5567 London homes, where each household's  electricity usage is recorded every half hour  \citep{ElectricityData}. The functional covariate $X(t)$ is defined as the 48 measurements per household that constitute one day's electricity usage curve. Based on this $X(t)$, we consider four  prediction tasks with different response variables: 
\\[-2em]
\begin{enumerate} \itemsep-0.1cm
\item[1.] Predict a household's 
total electricity consumption in the next week (week 2) based on its $X(t)$. [\emph{regression}] 
\item[2.] Predict a household's total electricity consumption in a later week (week 5) based on its $X(t)$.  [\emph{regression}] 
\item[3.] Predict whether a household's morning (6am-12pm) electricity consumption in week 5 exceeds a certain threshold based on its $X(t)$.  [\emph{classification}]
\item[4.] Predict whether a household's consumption during the day (8am-17pm) exceeds night usage (17pm-12am) by a threshold in week 5 based on its $X(t)$.  [\emph{classification}] 
\\[-2em]
\end{enumerate}
The threshold values in tasks 3 and 4 are selected to ensure approximate class-balance in these classification problems. Because household consumption behavior is  likely to change over a longer period, Tasks 2-4 are expected to be harder than Task 1. Results for these tasks are reported in columns Tasks 1-4 in Table~\ref{table2}.

\noindent {\bf Wearable Device Data:} This data consist of wearable device data from the National Health and Nutrition Examination Survey (NHANES) \citepalias{CDCData}. Each subject in the study wore a device that continuously  measures the intensity level of their physical activities within one week. The functional covariate $X(t)$ is the average activity levels every 30 minutes for one full day, resulting in a curve of 48 observations per subject. We examine whether physical activities  are predictive of various health outcomes: 
\\[-1em]

\begin{enumerate} \itemsep-0.1cm
\item[5.] Predict whether a subject has diabetes. [\emph{classification}]
\item[6.] Predict if subject feels chest pain. [\emph{classification}]
\item[7.] Predict whether a subject experiences shortness of breath on stairs. [\emph{classification}]
\\[-2em]
\end{enumerate}
Tasks 6 and 7 aim at predicting a subject's cardiovascular health. Results for Tasks 5-7 are reported in  column Task 5 to column Task 7 in Table~\ref{table2}.

\noindent {\bf Mexfly and Medfly Data:} 
The final two datasets pertain to Mexican fruit flies (Mexfly) \citep{CLMWSH2005} and Mediterranean fruit flies (Medfly) \citep{CMWC2003}, recording the number of eggs laid daily for each fly. Our task is to use early trajectories of egg-laying (daily number of eggs laid) to predict the \emph{lifetime reproduction}, defined as the total number of eggs laid by the fly over its lifetime:
\\[-2em]
\begin{enumerate} \itemsep-0.1cm
\item[8.] Predict lifetime reproduction of a Mexfly using   its egg-laying curve $X(t)$ from day 1 to 30. [\emph{regression}] 
\item[9.] Predict lifetime reproduction of a Medfly using   its egg-laying curve $X(t)$ from day 1 to 20. [\emph{regression}] 
\\[-2em]
\end{enumerate}
The choice of the thresholds, day 20 and 30, is motivated by predicting lifetime reproduction based on early reproduction pattern in pre-peak period (peak usually occurs after 20 or 30 days depending on the species). Results are presented in Table~\ref{table2} in columns Tasks 8 and 9.

\textbf{Results.} Empirically, AdaFNN performs better than all baseline methods in all 9 prediction tasks, demonstrating its advantage for diverse forms of real functional data spanning regression/classification problems. In contrast, none of the baseline methods consistently outperformed all other baselines across these tasks.  
Basis orthogonality and sparsity were used to improve learned representations and possibly get a better fit of the data (but like all regularization, the effectiveness of our proposed regularizers varies from dataset to dataset).
Many of the best reported results are from AdaFNN with penalty $\lambda_1 > 0$, demonstrating that our orthogonal regularization technique improves the learned functional representations. While $\lambda_2 > 0$ only produces the most accurate AdaFNN model for one of the tasks, the $L_1$ penalty can remain  useful for interpretability of the model. 
As with all regularizers, the optimal degree of regularization to employ also varies from dataset to dataset.  
By leveraging its superior representational capabilities, AdaFNN is also able to achieve superior accuracies with fewer bases than the B-spline + NN or FPCA + NN baselines.


\section{Conclusion} \label{future}
This work presents a new approach to adapt representation learning techniques for functional data. Our proposed architecture does not require handcrafted bases to handle functional inputs, and learns the optimal bases for a particular dataset in an end-to-end manner. The Basis Layer compresses functional covariates in a linear fashion into a low-dimensional vector that reflects only those factors of variation most relevant to the response value. Traditional dimension reduction techniques like FPCA instead attempt to capture all variation in the functional input itself, regardless of its relationship to the response. There are many disadvantages to retaining global information about $X(t)$ rather than merely what is needed to infer $Y$, some of which are outlined in the \emph{information bottleneck} principle of \citet{tishby2015deep}. 

Note that AdaFNN can be easily extended to vector-valued functional data, where either $t \in \mathbb{R}^p$ or $X(t) \in \mathbb{R}^p$ for $p > 1$. In the former case, each basis layer micro NN simply operates on vector-valued inputs $t$, while in the latter case, these micro NN would have larger output layers to produce vectors rather than scalars.
Furthermore, our method can also be applied to data with both multiple functional covariates (can simply employ separate basis layers for each and pool their outputs) as well as auxiliary vector covariates in addition to $X(t)$ (can simply concatenate our basis layer output $\tilde{v}_c$ with these additional covariates before it is fed into the subsequent feedforward network). 

As previously mentioned, AdaFNN is also directly applicable to non-uniform observations of $X(t)$, where the $t_j$ are not equally spaced, although such cases require proper selection of the integration weights $\omega_j$. 
However, successful application of AdaFNN to sparsely observed functional data, where the underlying process $X(t)$ is only observed at few locations $t_j$, remains nontrivial and likely requires improved numerical integration strategies as well as stronger inductive bias in the micro NN architecture \cite{gunter2014sampling}. 
Nonetheless, we expect our adaptive Basis Layer will find broad applicability as a general-purpose representation learning tool for domains with functional data such as wearable devices, climatology, genomics, or neuroimaging. 

\section{Acknowledgements}
We would like to thank the reviewers for their constructive feedback. The research of Jane-Ling Wang was supported by NSF grant  19-14917.

\clearpage
\bibliography{reference}

\begin{thebibliography}{35}
\providecommand{\natexlab}[1]{#1}
\providecommand{\url}[1]{\texttt{#1}}
\expandafter\ifx\csname urlstyle\endcsname\relax
  \providecommand{\doi}[1]{doi: #1}\else
  \providecommand{\doi}{doi: \begingroup \urlstyle{rm}\Url}\fi

\bibitem[Besse \& Ramsay(1986)Besse and Ramsay]{BR1986}
Besse, P. and Ramsay, J.~O.
\newblock Principal components analysis of sampled functions.
\newblock \emph{Psychometrika}, 51\penalty0 (2):\penalty0 285--311, 1986.

\bibitem[Cardot et~al.(1999)Cardot, Ferraty, and Sarda]{CFS1999}
Cardot, H., Ferraty, F., and Sarda, P.
\newblock Functional linear model.
\newblock \emph{Statistics \& Probability Letters}, 45\penalty0 (1):\penalty0
  11--22, 1999.

\bibitem[Cardot et~al.(2003{\natexlab{a}})Cardot, Ferraty, and Sarda]{CFS1991}
Cardot, H., Ferraty, F., and Sarda, P.
\newblock Spline estimators for the functional linear model.
\newblock \emph{Statistica Sinica}, pp.\  571--591, 2003{\natexlab{a}}.

\bibitem[Cardot et~al.(2003{\natexlab{b}})Cardot, Ferraty, and Sarda]{CFS2003}
Cardot, H., Ferraty, F., and Sarda, P.
\newblock Spline estimators for the functional linear model.
\newblock \emph{Statistica Sinica}, pp.\  571--591, 2003{\natexlab{b}}.

\bibitem[Carey et~al.(2005)Carey, Liedo, M{\"u}ller, Wang, Senturk, and
  Harshman]{CLMWSH2005}
Carey, J.~R., Liedo, P., M{\"u}ller, H.-G., Wang, J.-L., Senturk, D., and
  Harshman, L.
\newblock Biodemography of a long-lived tephritid: reproduction and longevity
  in a large cohort of female mexican fruit flies, anastrepha ludens.
\newblock \emph{Experimental Gerontology}, 40\penalty0 (10):\penalty0 793--800,
  2005.

\bibitem[Chiou(2012)]{C2012}
Chiou, J.-M.
\newblock Dynamical functional prediction and classification, with application
  to traffic flow prediction.
\newblock \emph{Annals of Applied Statistics}, 6\penalty0 (4):\penalty0
  1588--1614, 2012.

\bibitem[Chiou et~al.(2003)Chiou, M{\"u}ller, Wang, and Carey]{CMWC2003}
Chiou, J.-M., M{\"u}ller, H.-G., Wang, J.-L., and Carey, J.~R.
\newblock A functional multiplicative effects model for longitudinal data, with
  application to reproductive histories of female medflies.
\newblock \emph{Statistica Sinica}, 13\penalty0 (4):\penalty0 1119, 2003.

\bibitem[Dou et~al.(2012)Dou, Pollard, and Zhou]{DPZ2012}
Dou, W.~W., Pollard, D., and Zhou, H.~H.
\newblock Estimation in functional regression for general exponential families.
\newblock \emph{Annals of Statistics}, 40\penalty0 (5):\penalty0 2421--2451,
  2012.

\bibitem[Ferraty \& Vieu(2006)Ferraty and Vieu]{FV2006}
Ferraty, F. and Vieu, P.
\newblock \emph{Nonparametric Functional Data Analysis: Theory and Practice}.
\newblock Springer Science \& Business Media, 2006.

\bibitem[Gunter et~al.(2014)Gunter, Osborne, Garnett, Hennig, and
  Roberts]{gunter2014sampling}
Gunter, T., Osborne, M.~A., Garnett, R., Hennig, P., and Roberts, S.~J.
\newblock Sampling for inference in probabilistic models with fast bayesian
  quadrature.
\newblock In \emph{Advances in Neural Information Processing Systems}, 2014.

\bibitem[Guss(2016)]{G2016}
Guss, W.~H.
\newblock Deep function machines: Generalized neural networks for topological
  layer expression.
\newblock \emph{arXiv preprint arXiv:1612.04799}, 2016.

\bibitem[Guss \& Salakhutdinov(2019)Guss and Salakhutdinov]{GS2019}
Guss, W.~H. and Salakhutdinov, R.
\newblock On universal approximation by neural networks with uniform guarantees
  on approximation of infinite dimensional maps.
\newblock \emph{arXiv preprint arXiv:1910.01545}, 2019.

\bibitem[Hardt et~al.(2016)Hardt, Recht, and Singer]{H2016}
Hardt, M., Recht, B., and Singer, Y.
\newblock Train faster, generalize better: Stability of stochastic gradient
  descent.
\newblock In \emph{Proceedings of the 33rd International Conference on
  International Conference on Machine Learning}, volume~48, pp.\  1225--1234.
  JMLR, 2016.

\bibitem[Hsing \& Eubank(2015)Hsing and Eubank]{HE2015}
Hsing, T. and Eubank, R.
\newblock \emph{Theoretical Foundations of Functional Data Analysis, with an
  Introduction to Linear Operators}, volume 997.
\newblock John Wiley \& Sons, 2015.

\bibitem[James et~al.(2009)James, Wang, Zhu, et~al.]{JWZ09}
James, G.~M., Wang, J., Zhu, J., et~al.
\newblock Functional linear regression that's interpretable.
\newblock \emph{Annals of Statistics}, 37\penalty0 (5A):\penalty0 2083--2108,
  2009.

\bibitem[Leng \& M{\"u}ller(2006)Leng and M{\"u}ller]{LM2006}
Leng, X. and M{\"u}ller, H.-G.
\newblock Classification using functional data analysis for temporal gene
  expression data.
\newblock \emph{Bioinformatics}, 22\penalty0 (1):\penalty0 68--76, 2006.

\bibitem[Li et~al.(2010)Li, Hsing, et~al.]{LH2010}
Li, Y., Hsing, T., et~al.
\newblock Uniform convergence rates for nonparametric regression and principal
  component analysis in functional/longitudinal data.
\newblock \emph{Annals of Statistics}, 38\penalty0 (6):\penalty0 3321--3351,
  2010.

\bibitem[Lin et~al.(2013)Lin, Chen, and Yan]{LCY2013}
Lin, M., Chen, Q., and Yan, S.
\newblock Network in network.
\newblock \emph{arXiv preprint arXiv:1312.4400}, 2013.

\bibitem[M{\"u}ller(2005)]{M2005}
M{\"u}ller, H.-G.
\newblock Functional modelling and classification of longitudinal data.
\newblock \emph{Scandinavian Journal of Statistics}, 32\penalty0 (2):\penalty0
  223--240, 2005.

\bibitem[M{\"u}ller \& Stadtm{\"u}ller(2005)M{\"u}ller and
  Stadtm{\"u}ller]{MS2005}
M{\"u}ller, H.-G. and Stadtm{\"u}ller, U.
\newblock Generalized functional linear models.
\newblock \emph{Annals of Statistics}, 33\penalty0 (2):\penalty0 774--805,
  2005.

\bibitem[{National Center for Health Statistics (NCHS), Centers for Disease
  Control and Prevention (CDC)}(2020)]{CDCData}
{National Center for Health Statistics (NCHS), Centers for Disease Control and
  Prevention (CDC)}.
\newblock National health and nutrition examination survey data., 2020.

\bibitem[Ramsay \& Silverman(2007)Ramsay and Silverman]{R2007}
Ramsay, J.~O. and Silverman, B.~W.
\newblock \emph{Applied Functional Data Analysis: Methods and Case Studies}.
\newblock Springer, 2007.

\bibitem[Rice \& Silverman(1991)Rice and Silverman]{RS1991}
Rice, J.~A. and Silverman, B.~W.
\newblock Estimating the mean and covariance structure nonparametrically when
  the data are curves.
\newblock \emph{Journal of the Royal Statistical Society: Series B
  (Methodological)}, 53\penalty0 (1):\penalty0 233--243, 1991.

\bibitem[Rice \& Wu(2001)Rice and Wu]{RW2001}
Rice, J.~A. and Wu, C.~O.
\newblock Nonparametric mixed effects models for unequally sampled noisy
  curves.
\newblock \emph{Biometrics}, 57\penalty0 (1):\penalty0 253--259, 2001.

\bibitem[Rossi \& Conan-Guez(2005)Rossi and Conan-Guez]{RC2005}
Rossi, F. and Conan-Guez, B.
\newblock Functional multi-layer perceptron: a non-linear tool for functional
  data analysis.
\newblock \emph{Neural Networks}, 18\penalty0 (1):\penalty0 45--60, 2005.

\bibitem[Rossi et~al.(2002)Rossi, Conan-Guez, and Fleuret]{RCF2002}
Rossi, F., Conan-Guez, B., and Fleuret, F.
\newblock Functional data analysis with multi layer perceptrons.
\newblock In \emph{Proceedings of the 2002 International Joint Conference on
  Neural Networks.}, volume~3, pp.\  2843--2848. IEEE, 2002.

\bibitem[Rossi et~al.(2005)Rossi, Delannay, Conan-Guez, and
  Verleysen]{RDCV2005}
Rossi, F., Delannay, N., Conan-Guez, B., and Verleysen, M.
\newblock Representation of functional data in neural networks.
\newblock \emph{Neurocomputing}, 64:\penalty0 183--210, 2005.

\bibitem[Silverman(1996)]{S1991}
Silverman, B.~W.
\newblock Smoothed functional principal components analysis by choice of norm.
\newblock \emph{Annals of Statistics}, 24\penalty0 (1):\penalty0 1--24, 1996.

\bibitem[Song et~al.(2008)Song, Deng, Lee, and Kwon]{SDK2008}
Song, J.~J., Deng, W., Lee, H.-J., and Kwon, D.
\newblock Optimal classification for time-course gene expression data using
  functional data analysis.
\newblock \emph{Computational Biology and Chemistry}, 32\penalty0 (6):\penalty0
  426--432, 2008.

\bibitem[Tishby \& Zaslavsky(2015)Tishby and Zaslavsky]{tishby2015deep}
Tishby, N. and Zaslavsky, N.
\newblock Deep learning and the information bottleneck principle.
\newblock In \emph{IEEE Information Theory Workshop}, pp.\  1--5. IEEE, 2015.

\bibitem[{UK Power Networks}(2015)]{ElectricityData}
{UK Power Networks}.
\newblock Smartmeter energy consumption data in london households, 2015.

\bibitem[Wang et~al.(2016)Wang, Chiou, and M{\"u}ller]{W2016}
Wang, J.-L., Chiou, J.-M., and M{\"u}ller, H.-G.
\newblock Functional data analysis.
\newblock \emph{Annual Review of Statistics and Its Application}, 3:\penalty0
  257--295, 2016.

\bibitem[Wang et~al.(2021)Wang, Lu, Zhang, and Hahn]{WLZH21}
Wang, Q., Lu, Y., Zhang, X., and Hahn, J.
\newblock Region of interest selection for functional features.
\newblock \emph{Neurocomputing}, 422:\penalty0 235--244, 2021.

\bibitem[Yao et~al.(2005)Yao, M{\"u}ller, and Wang]{YMW2005a}
Yao, F., M{\"u}ller, H.-G., and Wang, J.-L.
\newblock Functional data analysis for sparse longitudinal data.
\newblock \emph{Journal of the American Statistical Association}, 100\penalty0
  (470):\penalty0 577--590, 2005.

\bibitem[Zhou et~al.(2013)Zhou, Wang, and Wang]{ZWW13}
Zhou, J., Wang, N.-Y., and Wang, N.
\newblock Functional linear model with zero-value coefficient function at
  sub-regions.
\newblock \emph{Statistica Sinica}, 23\penalty0 (1):\penalty0 25, 2013.

\end{thebibliography}
\bibliographystyle{icml2021}
\clearpage


\section*{Supplementary material}

# Supplementary Materials
## for
## Deep Learning for Functional Data Analysis with Adaptive Basis Layers

## 1 Proof of Theorem 1

*Proof.* Let $\tau_i : \mathbb{R}^q \to \mathbb{R}$ be a projection such that $\tau_i(v) = v_i$ for $v = (v_1, \ldots, v_q) \in \mathbb{R}^q$. It is straightforward to verify that $\tau_i$s are linear and continuous and $v = (\tau_1(v), \ldots, \tau_q(v))$. Using this property, we can write $g = (\tau_1 \circ g, \ldots, \tau_q \circ g)$ with each $\tau_i \circ g : \mathcal{C}([0, 1]) \to \mathbb{R}$ being a continuous linear functional. We further denote $c_i = \tau_i \circ g(f)$ and write $v_c = (c_1, \ldots, c_q)^\top$. By Riesz representation theorem, for each $i = 1, \ldots, q$, there exists $b_i \in \mathcal{L}_2([0, 1])$ such that $\tau_i \circ g(f) = \langle f, b_i \rangle$ for every $f \in \mathcal{C}([0, 1])$ and $\|\tau_i \circ g\|_{\text{op}} = \sup_{\|f\|=1} |\langle f, b_i \rangle| = \|b_i\|_2$. Since $\|f\|_2 \leq 1$, we can check that $\|v_c\|_2 \leq K$ where $K = \sqrt{q} \max_i \|b_i\|_2$.

Classical universal approximation results [Cybenko, 1989, Funahashi, 1989, Hornik, 1991, Stinchcombe, 1999] imply that for any $\epsilon > 0$, there exists a network with weights $\Theta^*$ such that $\sup_{\|v\|_2 \leq 2K} |\text{nn}_{\Theta^*}(v) - h(v)| < \epsilon/2$. Since $h$ is uniformly continuous on the compact set $D = \{v \in \mathbb{R}^q \mid \|v\|_2 \leq 2K\}$, there exists $0 < \rho_\epsilon < K$ such that for any $v_1, v_2 \in D$, $\|v_1 - v_2\|_2 < \rho_\epsilon$ implies $|h(v_1) - h(v_2)| < \epsilon/2$.

On the other hand, it is well known that $\mathcal{C}([0, 1])$ is dense in $\mathcal{L}_2([0, 1])$. For each $i = 1, \ldots, q$, there exists $\tilde{b}_i \in \mathcal{C}([0, 1])$ such that $\|\tilde{b}_i - b_i\|_2 < \rho_\epsilon/(2\sqrt{q})$. Classical universal approximation results imply that there exists a set of networks, each of which has weights $\Theta_i^*$, such that $\sup_{t \in [0,1]} |\text{nn}_{\Theta_i^*}(t) - \tilde{b}_i(t)| < \rho_\epsilon/(2\sqrt{q})$. Denote $\tilde{c}_i = \langle \text{nn}_{\Theta_i^*}, f \rangle$ and write $\tilde{v}_c = (\tilde{c}_1, \ldots, \tilde{c}_q)^\top$. Then, we have

$$|\tilde{c}_i - c_i| \leq |\langle \text{nn}_{\Theta_i^*} - \tilde{b}_i, f \rangle| + |\langle \tilde{b}_i - b_i, f \rangle| < \rho_\epsilon/\sqrt{q}$$

and thus $\|\tilde{v}_c - v_c\|_2 < \rho_\epsilon$.

Therefore, by linking these steps together, we have

$$\sup_{\substack{f \in \mathcal{C}([0,1]) \\ \|f\|_2 \leq 1}} |\widehat{\mathcal{T}}^*(f) - \mathcal{T}(f)| \leq \sup_{\substack{\|\tilde{v}_c - v_c\|_2 < \rho_\epsilon \\ v_c, \tilde{v}_c \in D}} |\text{nn}_{\Theta^*}(\tilde{v}_c) - h(v_c)|$$

$$\leq \sup_{v \in D} |\text{nn}_{\Theta^*}(v) - h(v)| + \sup_{\substack{\|v_1 - v_2\|_2 < \rho_\epsilon \\ v_1, v_2 \in D}} |h(v_1) - h(v_2)|$$

$$< \epsilon/2 + \epsilon/2 = \epsilon.$$

For the second part of the claim, first note that $X$ is a continuous random process defined on a compact interval. Its norm $\|X\|_2$ is a random variable on $\mathbb{R}$ and thus $\|X\|_2$ is stochastically bounded. In other words, for any $\delta > 0$, there exists a constant $M_\delta > 0$ such that $\mathbb{P}(\|X\|_2 \leq M_\delta) > 1 - \delta$. Then, it is easy to verify that by using $\|X\|_2 \leq M_\delta$ in replacement of $\|f\|_2 \leq 1$ in the arguments above, we can obtain $\sup_{f \in \mathcal{C}([0,1]), \|f\|_2 \leq M_\delta} |\widehat{\mathcal{T}}^\star(f) - \mathcal{T}(f)| < \delta$ for a suitable $\widehat{\mathcal{T}}^\star$. □

## 2 Proof of Theorem 2

*Proof.* Without loss of generality, we may assume that the model $\widehat{\mathcal{T}}_\Theta$ has one basis node in its BL and one hidden layer (with $m$ nodes) in the subsequent network. Suppose that a numerical integration algorithm with weights $\{\omega_j\}_{j=1}^{J+1}$ and grid $\{t_j\}_{j=1}^{J+1}$ is used. Then, the model $\widehat{\mathcal{T}}_\Theta$ can be expressed as $\widehat{\mathcal{T}}_\Theta(X) = \sigma\left(\sum_{l=1}^m \theta_{l1} \sigma\left(\theta_{l2} \sum_{j=1}^{J+1} \omega_j X(t_j) \cdot \text{nn}_{\tilde{\Theta}}(t_j) + \theta_{l3}\right) + \theta_4\right)$, where $\sigma$ is a nonlinear Lipschitz activation function (e.g., sigmoid or ReLU), and $\tilde{\Theta}$ denotes the parameters of the basis function network $\text{nn}_{\tilde{\Theta}}$. Note that here $\Theta$ denotes the collection of $\tilde{\Theta}$, $\theta_{l1}$s, $\theta_{l2}$s, $\theta_{l3}$s, and $\theta_4$.

It is straightforward to verify that both $\widehat{\mathcal{T}}_\Theta$ and $\nabla_\Theta \widehat{\mathcal{T}}_\Theta$ are bounded Lipschitz functions in $\Theta$ with bounded Lipschitz constants due to the condition $\sup_{t \in [0,1]} |X(t)| \leq M_1$ and Assumption (i). We may also assume that

the the loss function $\ell(\cdot,\cdot)$ is non-negative, since we can always shift the bounded loss function by a positive constant so that its minimum value is non-negative. Using the condition $|Y| \leq M_2$, Assumption (ii), chain rule, and the fact that a composition of two Lipschitz functions is also a Lipschitz function, we conclude that $\ell(\widehat{\mathcal{T}}_\Theta(X), Y)$ and $\nabla_\Theta \ell(\widehat{\mathcal{T}}_\Theta(X), Y)$ are also Lipschitz in $\Theta$ with bounded Lipschitz constants. Therefore, the second part of the theorem follows directly from Theorem 3.12 in Hardt et al. [2016] by checking its conditions. □

## 3 Experiment Details

Table 1 summarizes the number of bases used in each of the four simulation settings and each of the nine tasks in the data experiments.

Table 1: Number of bases used for the four simulation settings and nine tasks on datasets. The error rate of each method (already in the main paper) is also reported right below the method. The smallest error is marked in bold.

| Method | Case 1 | Case 2 | Case 3 | Case 4 | Task 1 | Task 2 | Task 3 | Task 4 | Task 5 | Task 6 | Task 7 | Task 8 | Task 9 |
|---|---|---|---|---|---|---|---|---|---|---|---|---|---|
| Raw data + NN | 51 | 51 | 51 | 51 | 48 | 48 | 48 | 48 | 48 | 48 | 48 | 30 | 20 |
|  | 0.015 | 0.038 | 0.275 | 0.334 | 0.099 | 0.284 | 0.124 | 0.296 | 0.380 | 0.488 | 0.472 | 0.406 | 0.373 |
| B-spline (4) + NN | 4 | 4 | 4 | 4 | 4 | 4 | 4 | 4 | 4 | 4 | 4 | 4 | 4 |
|  | 0.050 | 0.984 | 0.971 | 0.369 | - | - | - | - | - | - | - | - | - |
| B-spline (15) + NN | 15 | 15 | 15 | 15 | 15 | 15 | 15 | 15 | 15 | 15 | 15 | 15 | 15 |
|  | 0.013 | 0.019 | 0.206 | 0.251 | 0.094 | 0.306 | 0.137 | 0.326 | 0.335 | **0.477** | 0.429 | 0.413 | 0.387 |
| FPCA$_{0.9}$ + NN | 2 | 3 | 4 | 20 | 5 | 5 | 10 | 11 | 1 | 1 | 1 | 4 | 3 |
|  | 0.917 | 0.023 | 0.134 | 0.855 | - | - | - | - | - | - | - | - | - |
| FPCA$_{0.99}$ + NN | 4 | 19 | 20 | 28 | 17 | 17 | 24 | 24 | 5 | 4 | 2 | 12 | 7 |
|  | 0.003 | 0.036 | 0.239 | 0.667 | 0.119 | 0.339 | 0.143 | 0.306 | 0.363 | 0.493 | 0.431 | 0.429 | 0.378 |
| AdaFNN | 2 | 3 | 3 | 2 | 4 | 4 | 4 | 4 | 4 | 4 | 4 | 4 | 4 |
|  | **0.001** | **0.003** | **0.127** | **0.193** | **0.084** | **0.260** | **0.118** | **0.294** | **0.339** | 0.477 | **0.410** | **0.362** | **0.368** |

### 3.1 Model description

**Preprocessing before training.** For all tasks, the functional input is standardized entry-wise using function `StandardScaler` in `Python` package `sklearn.preprocessing`. The response is also standardized using the same function in all regression tasks.

**Basis Layer in AdaFNN.** As each basis node in the Basis Layer is implemented as a micro network, the scale of its output can vary with the initialization of the micro network weights. To stabilize the numerical integration at a basis node, we can normalize the output of the micro networks, i.e., scale the output of each basis node such that its numerical integral is equal to 1. Another benefit of this normalization is that we do not have to normalize the learned basis functions again when applying the orthogonality/sparsity regularization.

**B-spline scores.** The B-spline scores used to represent functional inputs as a baseline method are computed using the spline functions (e.g., `smooth.spline`) in `R`, and the coefficients are computed individually for each curve.

**FPCA scores.** We use the function `FPCA` in the `R` package `fdapace` [Chen et al., 2019] to compute functional principal component scores. Principal components are estimated and selected based on the training dataset first and then used to compute the principal component scores of curves in the test dataset.

### 3.2 Data description

**Simulated Datasets:** Simulation datasets are generated based on the following model

$$X(t) = \sum_{k=1}^{50} c_k \phi_k(t), \quad t \in [0,1],$$

where terms on the right hand are defined as:

1. $\phi_1(t) = 1$ and $\phi_k(t) = \sqrt{2}\cos((k-1)\pi t)$, $k = 2, \ldots, 50$;
2. $c_k = z_k r_k$, and $r_k$ are i.i.d. uniform random variables on $[-\sqrt{3}, \sqrt{3}]$.

Four simulation cases correspond to different configurations of $z_k$s:

1. In Case 1, $z_1 = 20, z_2 = z_3 = 5$, and $z_k = 1$ for $k \geq 4$. The response is $y = (\langle \phi_3, X \rangle)^2$;
2. In Case 2, $z_1 = z_3 = 5, z_5 = z_{10} = 3$, and $z_k = 1$ for other $k$. The response is $y = (\langle \phi_5, X \rangle)^2$.
3. Case 3 has the same configurations as Case 2. The observed response is
$$\tilde{y} = y + \epsilon = (\langle \phi_5, X \rangle)^2 + \epsilon, \quad \epsilon \sim N(0, 3/10).$$
For each time point $t_j$, the observed $X(t_j)$ is
$$\tilde{X}(t_j) = X(t_j) + \eta_j, \quad \eta_j \stackrel{\text{i.i.d.}}{\sim} N(0, 114/10).$$
4. In Case 4, $z_k = 1$ for all $k$. The response is $y = \langle \beta_2, X \rangle + (\langle \beta_1, X \rangle)^2$, where
$$\beta_1(t) = (4 - 16t) \cdot 1\{0 \leq t \leq 1/4\}$$
and
$$\beta_2(t) = (4 - 16|1/2 - t|) \cdot 1\{1/4 \leq t \leq 3/4\}.$$
The observed response is
$$\tilde{y} = y + \epsilon, \quad \epsilon \sim N(0, 1/10).$$
For each time point $t_j$, the observed $X(t_j)$ is
$$\tilde{X}(t_j) = X(t_j) + \eta_j, \quad \eta_j \stackrel{\text{i.i.d.}}{\sim} N(0, 5).$$
5. Case 5 has the same setup as Case 4, but with double the noise variance in $Y$.

In each case, 4000 curves are generated, among which 3200 are used for training while the rest 800 are left for testing. During training, 20% of the training data, i.e., 640 curves, are held out as a validation set.

Case 1 is designed to illustrate the weakness of the FPCA+NN approach, where the first two principle components explain at least 90% of the variability of the functional input $X$ and are selected, but the true signal $\phi_3$, which explains a very small fraction of the variability of $X$, is in the later principal components.

Case 2 is designed to illustrate the weakness of the B-spline+NN approach; here the true signal $\phi_5$ in the functional covariate cannot be well represented by a small set of, e.g., four, B-splines.

Case 3 is designed to illustrate that AdaFNN is able to learn meaning basis functions and does better than other baseline methods in the appearance of both measurement error and noise.

Case 4 is designed to illustrate that AdaFNN can be used to select relevant domains and achieve smaller prediction error.

Case 5 is designed to illustrate the effectiveness of the regularizers.

**Real Datasets:**

**Electricity Data [UK Power Networks, 2015].** The study contains electricity consumption readings for 5567 London households which participated in the Low Carbon London project from November 2011 to February 2014. Every half hour, there is a recording of the total electricity usage during the past half hour, so the total number of daily observations is 48. The observation periods vary among households. So, we select a period of 5 weeks in which the number of households in the project is the highest. This results in 5503 households. Since daily consumption is noisy, we take the average daily consumption during the first week period (to produce a smoother curve) as our functional covariate $X(t)$ for all prediction tasks. The training and test split is $4:1$, and 20% of the training data are held out for validation during the training process.

**NHANES Data [NCHS, CDC 2020].** The study contains wearable device readings of physical intensity of 7742 subjects during a one week period. According to the data documentation, the device was the ActiGraph AM-7164 (formerly the CSA/MTI AM-7164), manufactured by ActiGraph located in Ft. Walton Beach, FL. It is programmed to detect and record the magnitude of acceleration or "intensity" of movement. The original intensity readings were summed over 1-minute epoch. To construct the functional input, we further aggregate the readings over 30 minute intervals, and this results in 48 observations per day. We take the average daily intensity curve for the whole week as the functional input $X(t)$. The prediction tasks here is to classify the health conditions of individual subjects using their physical intensity curve $X(t)$. Since not everyone reported their health conditions and there are invalid and incomplete information in some subjects, we ended up with 6555 subjects for Task 5, 2416 for Task 6, and 2412 for Task 7. The training and test split is taken to be $4:1$, and 20% of training data are held out for validation during the training process.

**Mexfly [Carey et al., 2005] and Medfly[Chiou et al., 2003].** These two datasets are very similar, so we describe them together. The Mexfly data contain 1072 subjects and the Medfly data consist of 1000 subjects. For both data, the number of eggs laid daily was recorded for individual flies in the study until death. It is of biological interest to explore how early reproduction shapes lifetime reproduction, defined as the total number of eggs laid in a lifetime, which is perhaps the single most important biological question from the evolution point of view. We thus aim to predict the lifetime reproduction of a fly using its early reproduction trajectory $X(t)$, which is the number of eggs laid daily from birth till a specified day $M$ shortly before peak reproduction period. This day $M$ is 20 for the Medflies and 30 for the Mexflies. Next, we remove flies that died before day $M$ and the resulted sample size is 872 for Mexflies (Task 8) and 870 for Medflies (Task 9). The training and testing split $4:1$ is the same as before.

## Additional References for the Supplement

J. R. Carey, P. Liedo, H.-G. Müller, J.-L. Wang, D. Senturk, and L. Harshman. Biodemography of a long-lived tephritid: reproduction and longevity in a large cohort of female mexican fruit flies, anastrepha ludens. *Experimental Gerontology*, 40(10):793–800, 2005.

Y. Chen, C. Carroll, X. Dai, J. Fan, P. Z. Hadjipantelis, K. Han, H. Ji, H.-G. Mueller, and J.-L. Wang. *fdapace: Functional Data Analysis and Empirical Dynamics*, 2019. R package version 0.5.0.

J.-M. Chiou, H.-G. Müller, J.-L. Wang, and J. R. Carey. A functional multiplicative effects model for longitudinal data, with application to reproductive histories of female medflies. *Statistica Sinica*, 13(4): 1119, 2003.

G. Cybenko. Approximation by superpositions of a sigmoidal function. *Mathematics of Control, Signals and Systems*, 2(4):303–314, 1989.

K.-I. Funahashi. On the approximate realization of continuous mappings by neural networks. *Neural Networks*, 2(3):183–192, 1989.

M. Hardt, B. Recht, and Y. Singer. Train faster, generalize better: Stability of stochastic gradient descent. In *Proceedings of the 33rd International Conference on International Conference on Machine Learning*, volume 48, pages 1225–1234. JMLR, 2016.

K. Hornik. Approximation capabilities of multilayer feedforward networks. *Neural Networks*, 4(2):251–257, 1991.

National Center for Health Statistics (NCHS), Centers for Disease Control and Prevention (CDC). National health and nutrition examination survey data., 2020.

M. B. Stinchcombe. Neural network approximation of continuous functionals and continuous functions on compactifications. *Neural Networks*, 12(3):467–477, 1999.

UK Power Networks. Smartmeter energy consumption data in london households, 2015.



\end{document}